%% file: AAStemplatev2_0_6.tex
\pdfoutput=1

\documentclass[letterpaper, preprint, paper,11pt]{AAS}	
\usepackage{amsmath}
\usepackage{amssymb}
\usepackage{subcaption}
\usepackage{booktabs}
\usepackage{multirow}
\usepackage[numbers]{natbib} 

\input{macro}

\PaperNumber{24-213}

\begin{document}

\title{
Learning Optimal Control and Dynamical Structure of Global Trajectory Search Problems with Diffusion Models
}

\author{Jannik Graebner\thanks{PhD Student, Department of Mechanical and Aerospace Engineering, Princeton University, NJ, USA.},
Anjian Li\thanks{Ph.D. Candidate, Department of Electrical and Computer Engineering, Princeton University, NJ, USA},  
Amlan Sinha\thanks{Ph.D. Candidate, Department of Mechanical and Aerospace Engineering, Princeton University, NJ, USA}, \ and
Ryne Beeson\thanks{Assistant Professor, Department of Mechanical and Aerospace Engineering, Princeton University, NJ, USA}
}

\maketitle{}

\begin{abstract}
Spacecraft trajectory design is a global search problem, where previous work has revealed specific solution structures that can be captured with data-driven methods.
This paper explores two global search problems in the circular restricted three-body problem: hybrid cost function of minimum fuel/time-of-flight and transfers to energy-dependent invariant manifolds.
These problems display a fundamental structure either in the optimal control profile or the use of dynamical structures. 
We build on our prior generative machine learning framework to apply diffusion models to learn the conditional probability distribution of the search problem and analyze the model's capability to capture these structures.

\end{abstract}

\input{sections/intro}

\input{sections/problem_formulation}

\input{sections/cr3bp_problems}

\input{sections/methods}

\input{sections/results}

\input{sections/conclusion}

\section{Acknowledgment}
The presented simulations were performed on computational resources managed and supported by Princeton Research
Computing, a consortium of groups including the Princeton Institute for Computational Science and Engineering
(PICSciE) and the Office of Information Technology’s High-Performance Computing Center and Visualization
Laboratory at Princeton University. The authors would also like to acknowledge partial support for this effort from a Princeton University School of Engineering and Applied Sciences internal Seed Grant award. 


\bibliographystyle{AAS_publication}   

\input{AAStemplatev2_0_6.bbl}
\end{document}

%% file: macro.tex
\usepackage{bm}
\usepackage{amsmath}
\usepackage{subcaption}
\usepackage[colorlinks=true, pdfstartview=FitV, linkcolor=black, citecolor= black, urlcolor= black]{hyperref}
\usepackage{overcite}
\usepackage{footnpag}			      	
\usepackage{tikz}


%% file: sections/intro.tex
\section{Introduction}

Trajectory design for spacecraft is a parameterized global search problem.
To find trajectories that help spacecraft navigate within a complex multi-body gravitational environment, one needs to solve a highly non-convex optimal control problem with various problem parameters to be pre-specified, e.g. boundary conditions, control limits, etc.
We aim to find qualitatively different locally optimal solutions to this parameterized optimal control problem such that trade-offs can be made between design demands and constraints, and also allow more trajectory candidates in the early design phase to be ready for later use.

There has been a long history of hypotheses that the locally optimal solutions to the optimal control problems in spaceflight exhibit certain structures.
For example, the Pareto front was discovered in solutions to minimum time-of-flight and minimum fuel problems \cite{hartmann1998optimal, rauwolf1996near, russell2007primer, oshima2017global}.
Moreover, locally optimal solutions are assumed to be grouped into different clusters \cite{vasile2006preliminary, yam2011low, englander2012automated}, where many similar local optima are located within a single basin.
This assumption is demonstrated through the superior performance in efficient spacecraft trajectory design \cite{yam2011low, addis2011global, englander2017automated, vasile2008testing, izzo2010global, vasile2010analysis} using a global search algorithm Monotonic Basin Hopping (MBH) \cite{wales1997global, leary2000global}.
The MBH exploits the clustering structure by first conducting a global search using some heuristic distributions to find initial guesses, i.e. Cauchy or Pareto distribution, and then performing a local search using a gradient-based optimization solver to uncover the local optima nearby.
However, in the aforementioned work, there is limited visual demonstration of the clustering structure in the solutions of trajectory design problems.
In addition, the MBH, which mostly relies on heuristics for the search process, doesn't learn and exploit the solution structure in the problem but just always starts the global search from scratch.

\begin{figure}
    \centering
    \includegraphics[width=1.0\textwidth]{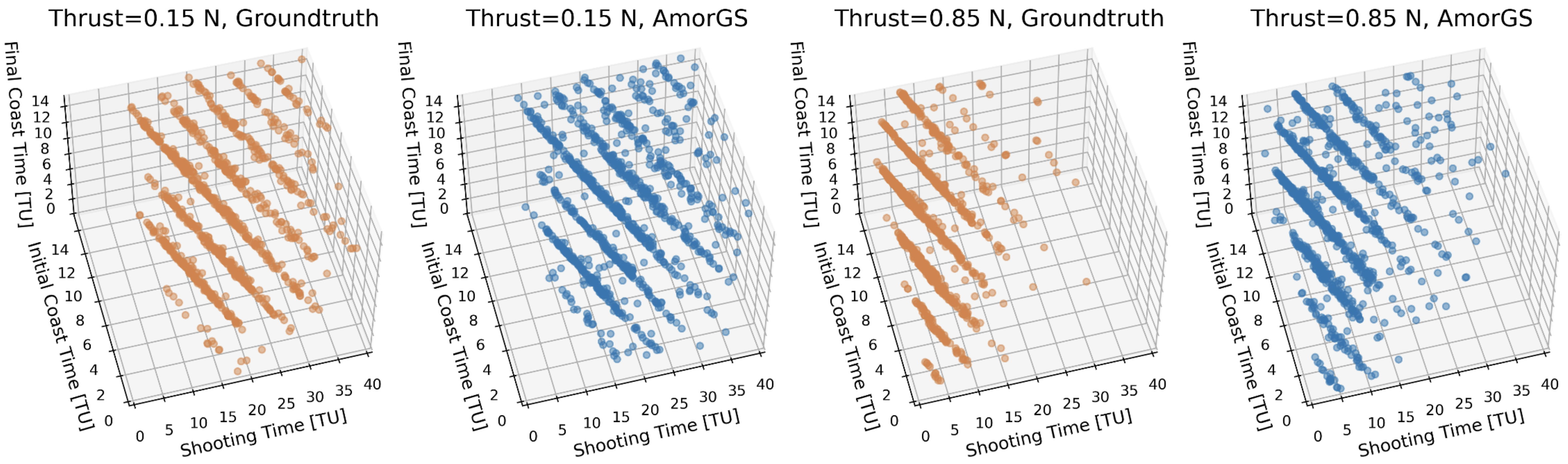}
    \caption{Hyperplane structure existed in locally optimal solutions to a cislunar transfer in CR3BP \cite{li2023amortized}.
    The initial/final coast time and shooting time of the spacecraft are presented in orange (Groundtruth solution) and blue (AmorGS prediction).
    The maximum allowable thrust $0.15$N and $0.85$N are unseen in the AmorGS training dataset.}
    \label{fig:cr3bp hyperplane}
\end{figure}

Recently, Li et al. \cite{li2023amortized, li2024efficient} studied a minimum-fuel low-thrust transfer in the Circular Restricted Three-Body Problem (CR3BP), where the trajectory starts from the end of a geostationary transfer orbit (GTO) and ends at a stable invariant manifold arc of a halo orbit around $\mathcal{L}_1$ Lagrange point.
In this global search problem, locally optimal solutions are found to have a clustering structure.
As shown in Figure \ref{fig:cr3bp hyperplane}, the ground truth solutions of time variables (orange), i.e. initial/final coast time and shooting time, are grouped into parallel hyperplanes, where the trajectory in each plane has similar time-of-flight.
Moreover, this hyperplane structure also varies with respect to the changes to the maximum allowable thrust of the spacecraft from $0.15$N to $0.85$N presented in Figure \ref{fig:cr3bp hyperplane}. 
To learn and leverage this discovered solution structure to provide a better initial guess for the optimal control problem, the \textbf{Amor}tized \textbf{G}lobal \textbf{S}earch (AmorGS) framework is presented \cite{li2023amortized}: With many parameterized optimal control problems solved offline, a conditional generative model is used to learn the clustering structure of the obtained solutions in order to predict the approximate solutions for the similar but unseen problems.
As presented in Figure \ref{fig:cr3bp hyperplane}, the AmorGS predictions (blue) can capture the hyperplane structure in the solution and make accurate predictions to unknown thrust $0.15$N and $0.85$N.
With these AmorGS predictions as initial guesses, the numerical optimization solver converges quickly and the computational efficiency of the online global search is largely improved.

Within AmorGS, Conditional Variational Autoencoder (CVAE) \cite{kingma2013auto, Sohn2015learning} as a popular generative model can be used to sample part of the solution variables (time variables) with hyperplane structure conditioned on the maximum allowable thrust. 
Together with Long-Short Term Memory (LSTM) \cite{hochreiter1997long}, it generates full-dimensional initial guess for optimal control problems with unseen maximum allowable thrust. \cite{li2023amortized}.
Furthermore, state-of-the-art diffusion models \cite{sohl2015deep, song2020score, ho2020denoising, ho2022classifier} are shown to be able to directly model the full-dimensional solution distribution \cite{li2024efficient} and automatically capture the complex structure in solutions without any human effort for structure identification.
It generates high-quality initial guesses for the problems that require much less time to converge with an optimization solver compared to CVAE and uniform sampling \cite{li2024efficient}.
But previous work \cite{li2023amortized, li2024efficient} only studies maximum allowable thrust as the varying parameter of the problems.

\subsection{Our Contributions}
In this paper, we further explore the optimal control structure and the use of dynamical structures in the solutions of two global search problems in the CR3BP, respectively.
For the first problem, we study a hybrid objective function that is a convex combination of minimum fuel and minimum time-of-flight for a cislunar transfer.
The problem parameter to vary is the weight of these two objectives.
We demonstrate that as we increase the weight of the minimum fuel objective, the locally optimal solutions tend to their theoretical bang-bang profile in the thrust magnitude based on Pontryagin's maximum principle \cite{doe2005}.

For the second problem, we study a cislunar transfer in the CR3BP with an energy-dependent halo orbit and its stable invariant manifold as a variable terminal boundary condition. 
In the previous work \cite{li2023amortized, li2024efficient}, the final boundary of the transfer is a fixed point on the stable invariant manifold arc of a halo orbit around $\mathcal{L}_1$ Lagrange point.
In this problem, we choose the varying problem parameter to be the energy level of the halo orbit and use a free terminal boundary condition on the invariant manifold.
We show that in the minimum-fuel transfer, the locally optimal solutions exhibit basin structures with respect to the trajectory endpoint on the manifold which depend on the energy level of the halo orbit.

Furthermore, we show that those optimal control structures and the use of dynamical structures in the solutions can be learned through the AmorGS framework and diffusion models.
We collect training data by solving two optimal control problems with various problem parameter values, i.e. different weights of the objectives and different energy levels of the target halo orbit. 
Then a diffusion model is trained on the collected data and learns how to sample optimal control solutions as initial guess conditioned on the objective weights or target halo energy.
Within this data-driven fashion, the solution structure, such as the bang-bang profile or the basin structure in the manifold parameters, are captured by the diffusion model and generalized to problems with unseen parameter values.
Since those structured samples from diffusion models have a similar distribution as the ground truth solution, many of them take very little time to converge to feasible or even locally optimal solutions.

%% file: sections/problem_formulation.tex
\section{Problem Formulation}
\subsection{Parameterized Optimal Control Problem}
The optimal control problem we want to solve is described by a cost function $J$ with problem parameter $\alpha$. The generic form is given by:
\begin{align}
    \label{equation: cost function}
    \min J(u; \alpha) \equiv \phi(\xi(t_f), t_f; \alpha) + \int_{t_0}^{t_f} \mathcal{L}(\xi_s, u_s, s ; \alpha) ds \nonumber \\
    s.t. \quad u \in \mathcal{U}, \ \textrm{and Eqs.} \ \eqref{equation: evolution differential equation}, \eqref{equation: initial and final boundary conditions}, \eqref{equation: path constraints} \ \textrm{are satisfied}, 
\end{align}
where $\xi_s$ represents the state of the spacecraft at timestep $s\in [t_0,t_f]$ with $t_0$ denoting the initial time and $t_f$ the final time. The state evolves according to the dynamic constraint
\begin{align}
    \label{equation: evolution differential equation}
    \xi_t = \xi_0 + \int_{t_0}^t f(\xi_s, s; \alpha) ds + \int_{t_0}^t g(\xi_s, u_s, s; \alpha) ds, \quad \forall t \in [t_0, t_f],
\end{align}
and satisfies the initial and terminal boundary conditions, 
\begin{align}
    \label{equation: initial and final boundary conditions}
    \xi_0 \equiv \xi(t_0) \in \Xi_0 \subseteq \mathbb{R}^{m} \quad \text{and} \quad \xi(t_f) \in \Xi_f \subseteq \mathbb{R}^{m}, 
\end{align}
as well as a set of path constraints, 
\begin{align}
    \label{equation: path constraints}
    \psi_k(\xi_s, s; \alpha) \leq 0, \quad \forall s \in [t_0, t_f], \quad \forall k \in \mathcal{K}. 
\end{align}

The total cost comprises a terminal cost $\phi$ and the running cost $\mathcal{L}$. The control $u$ is chosen from the set of admissible controls $\mathcal{U}$. We introduce the dynamics of the system with two vector fields. The initial vector field $f$ characterizes the inherent, natural dynamics of the system and the second part $g$ consists of any perturbing effects, including those exerted by our control. The sets of admissible initial and terminal boundary conditions are represented by $\Xi_0$ and $\Xi_f$, both required to be smooth submanifolds of $\mathbb{R}^{m}$. We use $\mathcal{K}$ to denote an index set for the path constraints with index $k$.

\subsection{Nonlinear Program}

We use a direct control transcription to convert the continuous-time optimal control problem into a nonlinear program (NLP), using a forward-backward shooting method \cite{betts1998survey}. 
The discretization of the state and control vectors in Eqs. \eqref{equation: cost function}, \eqref{equation: evolution differential equation}, \eqref{equation: initial and final boundary conditions} and \eqref{equation: path constraints}, yields a NLP of the form:
\begin{equation}
    \label{equation: nonlinear programming problem}
    \begin{split}
    &\quad \inf \limits_{x \in \mathcal{X}} J(x; \alpha), \\
    \text{subject to} &\quad c_i(x; \alpha) = 0, \quad \forall i \in \mathcal{E}, \\
    &\quad c_i(x; \alpha) \leq 0, \quad \forall i \in \mathcal{I},
    \end{split}
\end{equation}
where $\alpha$ represents the problem parameter. The cost function $J\in C^2$, the equality constraints $(c_i)_{i\in\mathcal{E}}\in C^1$ and the inequality constraints $(c_i)_{i\in\mathcal{I}}\in C^1$ are assumed to be nonlinear. We denote the index sets for equality and inequality constraints by $\mathcal{E}$ and $\mathcal{I}$. The discretized decision variable $x$ is chosen from a discretization $\mathcal{X}$ of the admissible control set $\mathcal{U}$.

A gradient-based numerical optimizer $\pi$ can be adopted to solve the NLP in Eq. \eqref{equation: nonlinear programming problem} to derive a solution when an initial guess of $x$ and parameter $\alpha$ is provided.
In this work, the sequential quadratic programming software SNOPT (short for Sparse Nonlinear OPTimizer) \cite{Gill.2005} is utilized. 
When the problem in Eq. \eqref{equation: nonlinear programming problem} is highly non-convex or high-dimensional, the global search for locally optimal solutions can be time-consuming due to high sample complexity and extensive solving time.

%% file: sections/cr3bp_problems.tex
\section{Circular Restricted Three Body Problem (CR3BP)}
\subsection{Equations of Motions}
The Circular Restricted Three Body Problem (CR3BP) is a dynamical model governing the motion of a point mass in a three-body gravitational system. It provides a good first-order approximation for the complex dynamical system and allows leveraging dynamical structures in preliminary mission design. The model assumes that a primary body with mass $m_1$ and a secondary body with mass $m_2<m_1$ follow a circular orbit around their barycenter. The orbital period of either body around the barycenter is denoted by $T$. The third body is a spacecraft of negligible mass. We normalize the spacecraft states using $(m_1+m_2)$ for mass, the constant distance between primary and secondary for length and $T/2\pi$ for time units. This reduces the parameters of the model to the gravitational constant $\mu = m_2/(m_1+m_2)$. The natural dynamics in this model are described in a canonical rotating frame with the primary body at $(-\mu,0,0)$ and the secondary body at $(1-\mu,0,0)$:
\begin{equation}
    \label{equation: dynamical system}
    \begin{split}
    &\ddot{q}_1 = 2\dot{q}_2+q_1-(1-\mu)\frac{q_1+\mu}{\rho^3_1}-\mu\frac{q_1-1+\mu}{\rho^3_2},\\
    &\ddot{q}_2 = -2\dot{q}_1+q_2-(1-\mu)\frac{q_2}{\rho^3_1}-\mu\frac{q_2}{\rho^3_2}, \\
    &\ddot{q}_3 = -(1-\mu)\frac{q_3}{\rho^3_1}-\mu\frac{q_3}{\rho^3_2},
    \end{split}
\end{equation}
the vectors $q$ and $\dot{q}$ describing position and velocity of the spacecraft. The scalars $\rho_1$ and $\rho_2$ represent the distances of the spacecraft to the primary and secondary body: 
\begin{align}
\rho_1 = \sqrt{(q_1+\mu)^2+q_2^2+q_3^2}, \quad \rho_2 = \sqrt{(q_1-1+\mu)^2+q_2^2+q_3^2}
\end{align}
The equilibrium points of the dynamical system described in Eq. \eqref{equation: dynamical system} are the five libration or Lagrange points. The first three, denoted by $\mathcal{L}_1$, $\mathcal{L}_2$ and $\mathcal{L}_3$ are unstable and colinear, located on the axes connecting the primary and secondary body. The other two equilateral points $\mathcal{L}_4$ and $\mathcal{L}_5$ are stable. The points $\mathcal{L}_1$ and $\mathcal{L}_2$, located closest to the secondary body, are of special interest for the trajectory design of space missions, since there are families of periodic orbits surrounding these points.
A libration point orbit that has become a popular location for space missions is the halo orbit, which is a periodic orbit with out-of-plane components around  $\mathcal{L}_1$, $\mathcal{L}_2$ or $\mathcal{L}_3$. \\
\ \\
The realms of possible motion of a particle in the CR3BP are dependent on its energy in relation to the energy levels at the libration points. This restricts particles with low energy from moving in between the interior or $m_1$ realm around the primary body, the smaller $m_2$ realm around the secondary body and the exterior realm. The periodic halo orbits possess stable and unstable invariant manifolds that connect the different realms of the CR3BP. Invariant manifolds transport material between the different realms of the CR3BP but can also be used to construct low-energy spacecraft trajectories. The energy level $e$ of the states on a halo orbit and its invariant manifolds is constant and depends on the altitude of the orbit. 

\subsection{Cislunar Mission}

We consider a continuous low-thrust problem.
The change in mass $\Dot{m}$ of the spacecraft can be described by the differential equation $\Dot{m} = \frac{|u|}{I_{sp}g}$
with the gravitational constant of earth $g \thickapprox 9.80665 \mathrm{m}/\mathrm{s}^2$. The specific impulse $I_{sp}$ is assumed to be constant and the absolute value of the control vector $|u|$ is equivalent to the thrust magnitude. We neglect other perturbations on the spacecraft such as solar pressure or atmospheric drag, making the vector field $g$ in Eq. \eqref{equation: evolution differential equation} only consist of the perturbation on $\ddot{q}$, described by the vector $u/m$. 
The spacecraft considered for the trajectory optimization problem has a total initial mass of $m_i =1000$ kg, consisting of $700$ kg fuel mass and $300$ kg dry mass. The electric propulsion system provides a constant specific impulse of $I_{sp}=1000$ s and thrust $T = 1$ N, resulting in a thrust acceleration of $0.001$ $\mathrm{m}/\mathrm{s}^2$.

The particular problem we investigate is a low thrust transfer in the Earth-Moon CR3BP, leveraging dynamical structures. The cislunar domain is gaining significance, with numerous missions planned by both governmental and private entities in the imminent future. Our starting orbit is a geostationary transfer orbit (GTO), a widely accessible orbit for a diverse range of launch systems. As the initial boundary for the low thrust trajectory optimization problem, we use the end of a fixed-time tangential low thrust spiral originating at the GTO. The target orbit is a halo orbit around the libration point $\mathcal{L}_1$. The spacecraft attains this halo orbit via coasting on its stable invariant manifold. Hence, the final boundary of the trajectory is on a specific arc on the invariant manifold. We do not consider any path constraints for this problem. 

In this paper, we consider two different example problems in the cislunar mission: Hybrid Cost Function and Variable Terminal Boundary Condition.
We aim to explore and exploit the solution structure in the global search of these two tasks.

\subsection{Example Problem I: Hybrid Cost Function}

In general, a space mission designer always has the goal to maximize payload and hence minimize fuel mass on board of the spacecraft. Additionally, a short time of flight is usually desirable, leading to reduced mission cost, resource optimization and risk mitigation. To allow a flexible weighting of those factors, we seek to minimize a hybrid cost function given by 
\begin{equation}
    \label{equation: hybrid cost function}
    J(u;\omega) = \omega\int_{t_0}^{t_f}-\frac{\dot{m}_s}{m_i}ds + (1-\omega)\int_{t_0}^{t_f}\frac{1}{t_{f,max}}dt
\end{equation}
displaying a convex combination of the amount of fuel consumed and the total time of flight. The parameter $\omega\in[0,1]$ specifies the weighting of the two terms, with $\omega = 0$ representing a minimum time transfer and $\omega = 1$ a minimum fuel transfer.
Both terms are normalized using the initial mass $m_i$ and the time of flight limit $t_{f,max}$, to ensure they are of the same order of magnitude.
We use a direct control transcription to discretize the decision variable for the optimal control problem described in Eq. \eqref{equation: nonlinear programming problem} resulting in the vector\cite{Beeson:2022aas}
\begin{equation}
    \label{equation: decision vector}
    x = (\tau_s,\tau_i,\tau_f,m_f,u_1,u_2,...,u_N)
\end{equation}
with $x\in\mathbb{R}^{3N+4}$. The total time of flight consists of the shooting time $\tau_s$ in addition to the initial and final coasting times $\tau_i$ and $\tau_f$. The shooting period is further divided into $N=20$ segments, with $u_i\in\mathcal{U}\subseteq\mathbb{R}^3$ describing the constant thrust vector during segment $i=1,2,...,N$ and $m_f$ being the final mass at the end of the mission. We can directly describe the cost function in terms of the components of the decision vector as 
\begin{equation}
    \label{equation: discrete hybrid cost function}
    J(u;\omega) = \omega\frac{-{m}_f}{m_i} + (1-\omega)\frac{(\tau_s+\tau_i+\tau_f)}{(\tau_{s,max}+\tau_{i,max}+\tau_{f,max})},
\end{equation}
with the minimization of the negative final mass being equivalent to minimizing the consumed fuel. 

For this problem, we target a halo orbit with perturbation energy $e_{pert}=0.01$ and orbital period $T^{\mathcal{H}}=2.748$.
The endpoint of a manifold arc defined through the parameters $t_1=0.2T^{\mathcal{H}}$ and $t_2=8$ serves as a fixed terminal boundary condition.
The temporal variables $t_1$ and $t_2$ are defined in Eq. \eqref{equation: state relation on manifold}.

In the case that $\omega = 1$, we know that the continuous-time locally optimal control solution must have a bang-bang profile for the thrust magnitude, and in the case of $\omega = 0$ the locally optimal control profiles will have maximal thrust magnitude for the entire time of flight. 
In this problem, we consider our conditional parameter $\alpha$ from Eq. \eqref{equation: cost function} to be equal to $\omega$ and aim to learn how the structure in the time variables, final mass and throttle profile changes when $\alpha = \omega$ is varied in the interval $[0, 1]$.

\subsection{Example Problem II: Variable Terminal Boundary Condition}

For the second problem, we study a transfer to an energy-dependent invariant manifold which is a variable terminal boundary condition.
Our objective is to analyze the basin structure of locally optimal solutions on the manifold and how this structure changes with halo orbits of different energy levels. 
We focus on solving a minimum fuel transfer, which is equivalent to setting $\omega=1$ in Eq. \eqref{equation: discrete hybrid cost function}. 
Two additional time variables $t_1$ and $t_2$ are added to the decision vector $x$ of the NLP to describe the terminal condition on the invariant halo manifold. 
With them, we extend the vector $x$ from Eq. \eqref{equation: decision vector} to be
\begin{equation}
    \label{equation: extended decision vector}
    x = (\tau_s,\tau_i,\tau_f,t_1,t_2,m_f,u_1,u_2,...,u_N),
\end{equation}
with $x\in\mathbb{R}^{3N+6}$. 
The time $t_1 \in [0, T^{\mathcal{H}})$ determines the location of the target manifold arc by specifying the insertion point of the backward integrated arc on the halo orbit, with $T^{\mathcal{H}}$ denoting the orbital period of the halo orbit. 
The second time parameter $t_2 \in[t_{2,min},t_{2,max}]$, with $t_{2,min}=5$ and $t_{2,max}=11$, is the time of flight on the manifold. 
The decision variables $t_1$ and $t_2$ are jointly optimized for the terminal state of the trajectory $\xi(T)$ at final time $T$, which is unambiguously described by 
\begin{equation}
    \label{equation: state relation on manifold}
    \begin{split}
    &\xi(T+t_2) = \xi_0^{\mathcal{H}(e)}+\int_{t_0^{\mathcal{H}(e)}}^{t_0^{\mathcal{H}(e)}+t_1}(f(\xi^{\mathcal{H}(e)}(\tau))+\epsilon_{\mathcal{H}(e)}(\tau)\delta(\tau-(t_0^{\mathcal{H}(e)}+t_1)))d\tau \\
    &\xi(T) = \xi(T+t_2)+\int_{T+t_2}^{T}f(\xi(\tau))d\tau, 
    \end{split}
\end{equation}
with the notation $(\cdot)^{\mathcal{H}(e)}$ indicating time and states on the halo orbit $\mathcal{H}$ with energy $e$. 
The first part of Equation \eqref{equation: state relation on manifold} describes the state of the spacecraft at the insertion point of the manifold arc on the halo orbit $\xi(T+t_2)$.
The state $\xi_0^{\mathcal{H}}$ represents the initial state on the halo orbit, which is defined as the intersection point of the halo orbit with the $q_1q_3$-plane of the CR3BP, located closest to earth $\xi_0^{\mathcal{H}} \equiv \xi^{\mathcal{H}}(t_0^{\mathcal{H}})=\{(\xi^{\mathcal{H}}_1,\xi^{\mathcal{H}}_2,\xi^{\mathcal{H}}_3)\,|\,\xi^{\mathcal{H}}_2=0,\,\xi_1^{\mathcal{H}}<\xi_{1}^{\mathcal{L}_1}\}$, with $\xi_{1}^{\mathcal{L}_1}$ denoting the $\xi_{1}$-coordinate of libration point $\mathcal{L}_1$.
We add the state evolution on the halo orbit during the time $t_1$ to that state, including $\epsilon_H$, which is a small perturbation aligned with the eigenvector associated with the negative eigenvalue of the monodromy matrix characterizing $\mathcal{H}$. 
This perturbation is added to the state after coasting on the halo orbit for $t_1$. 
To formulate this mathematically the dirac delta function $\delta(\tau-(t_0^{\mathcal{H}(e)}+t_1)))$ is used, with $\delta(\tau-(t_0^{\mathcal{H}(e)}+t_1)))=\infty$ for $\tau=t_0^{\mathcal{H}(e)}+t_1$ and $\delta(\tau-(t_0^{\mathcal{H}(e)}+t_1)))=0$ otherwise.
We then define $\xi_T$ in the second part of \eqref{equation: state relation on manifold} by backwards integrating on the manifold arc for the time period $t_2$, starting from $\xi(T+t_2)$.

\begin{figure}[!htbp]
    \centering
    \begin{tikzpicture}
        \node[anchor=south west,inner sep=0] (image) at (0,0) {\includegraphics[width=0.5\textwidth]{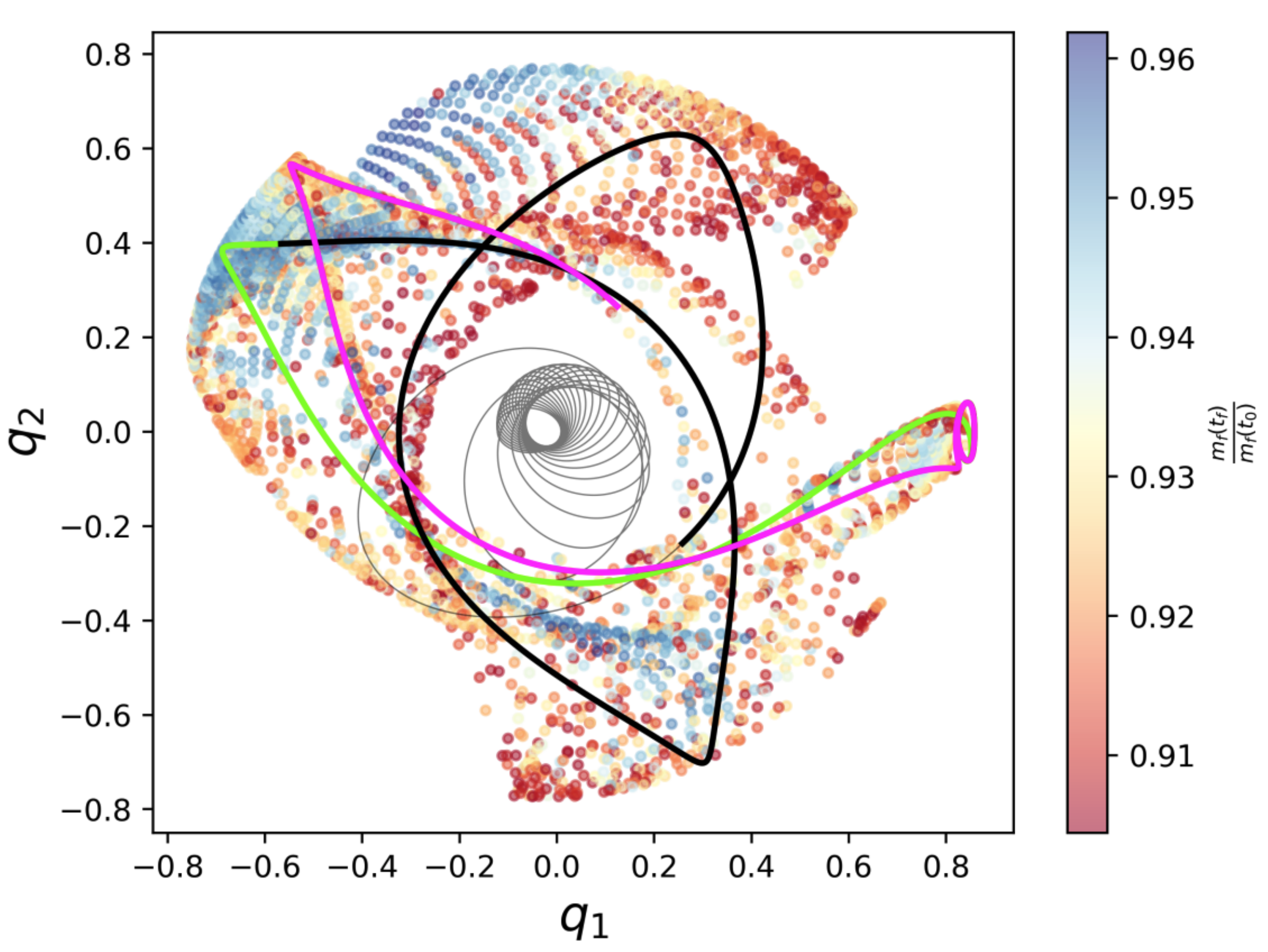}};
        \begin{scope}[x={(image.south east)},y={(image.north west)}]
            \fill[blue,opacity=0.15] (0.15,0.65) rectangle (0.3,0.8);
            \draw[blue, thick] (0.15,0.65) rectangle (0.3,0.8);
            \fill[blue,opacity=0.15] (0.25,0.8) rectangle (0.4,0.9);
            \draw[blue, thick] (0.25,0.8) rectangle (0.4,0.9);
            \fill[blue,opacity=0.15] (0.4,0.25) rectangle (0.55,0.35);
            \draw[blue, thick] (0.4,0.25) rectangle (0.55,0.35);
        \end{scope}
    \end{tikzpicture}
    \caption{Projection onto the Earth-Moon orbital plane of the GTO spiral (gray line), low thrust trajectory solution (black), the initial manifold arc that was targeted (magenta), and the final manifold arc that is targeted (green). 
    Optimal solutions to a discretized grid of the stable manifold in the variable space \(t_1\) and \(t_2\) are shown in the background, illustrating how the initial guess solution evolves to an insertion region of the invariant manifold that minimizes fuel usage, revealing multiple basins in the invariant manifolds.
    Examples of the basins are highlighted in blue.}
    \label{fig:cr3bp basin structure}
\end{figure}
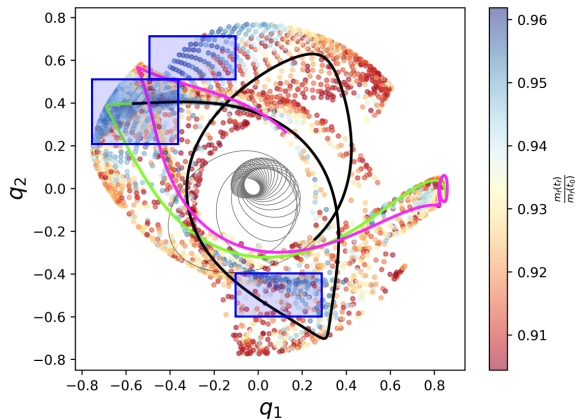

As a motivating example, consider a halo orbit $\mathcal{H}$ possessing an orbital energy $e$.
As mentioned above, the stable manifold arc $\mathcal{W}_S^\mathcal{H}(t_1, t_2)$ for the orbit can be characterized by the temporal parameters $t_1 \in [0, T^{\mathcal{H}})$ and $t_2 \in[0,t_{2,max}]$ for a specific $t_{2,max}$.
To facilitate an examination of the solution space with respect to the temporal parameters, the manifold arc can be discretized over a grid:
\begin{equation*}
    (t_1 \times t_2) = \{(t_i, t_j) \ | \ t_i \in t_1 \in [0, T^{\mathcal{H}}), t_j \in t_2 \in[0,t_{2,max}] \}.
\end{equation*}
By maintaining the same initial boundary condition and varying the final boundary condition across points in the grid, we solve for locally optimal trajectories within the solution space. 
This unveils a complex structure of multiple solution basins, as depicted in Figure \ref{fig:cr3bp basin structure}. 
Such basins in the solution space are indicative of an underlying structure in the problem, which can be leveraged for efficient global search algorithms.

Therefore, the $\alpha$ parameter of this problem describes the energy level of the halo orbit, and thus defines the invariant manifold that the terminal boundary lies on.
The energy level $e$ of the halo orbit is defined by the perturbation added to the energy of the libration point $\mathcal{L}_1$:
\begin{equation}
    e = e_{\mathcal{L}_1} + e_{pert,min} + \alpha (e_{pert,max}-e_{pert,min}),
    \label{equation: halo energy}
\end{equation}
with the parameter $\alpha \in [0,1]$.
The energy of $\mathcal{L}_1$ is given as $e_{\mathcal{L}_1}=-1.594$.
We vary the perturbation energy within the bounds of $e\in [e_{min},e_{max}]$, in a range where no bifurcations occur. 
This allows us to observe how the basin structure of the locally optimal solutions on the manifold change with respect to the halo orbits of different energy levels. 
We use a diffusion model to capture this basin structure and generalize to unseen terminal boundaries defined by the new energy level of the halo orbit.

%% file: sections/methods.tex
\section{Methodology}

In this section, we introduce the Amortized Global Search (AmorGS) framework \cite{li2023amortized} and the diffusion model details.

\subsection{AmorGS framework}

\begin{figure}[!htbp]
    \centering
    \includegraphics[width=0.9\textwidth]{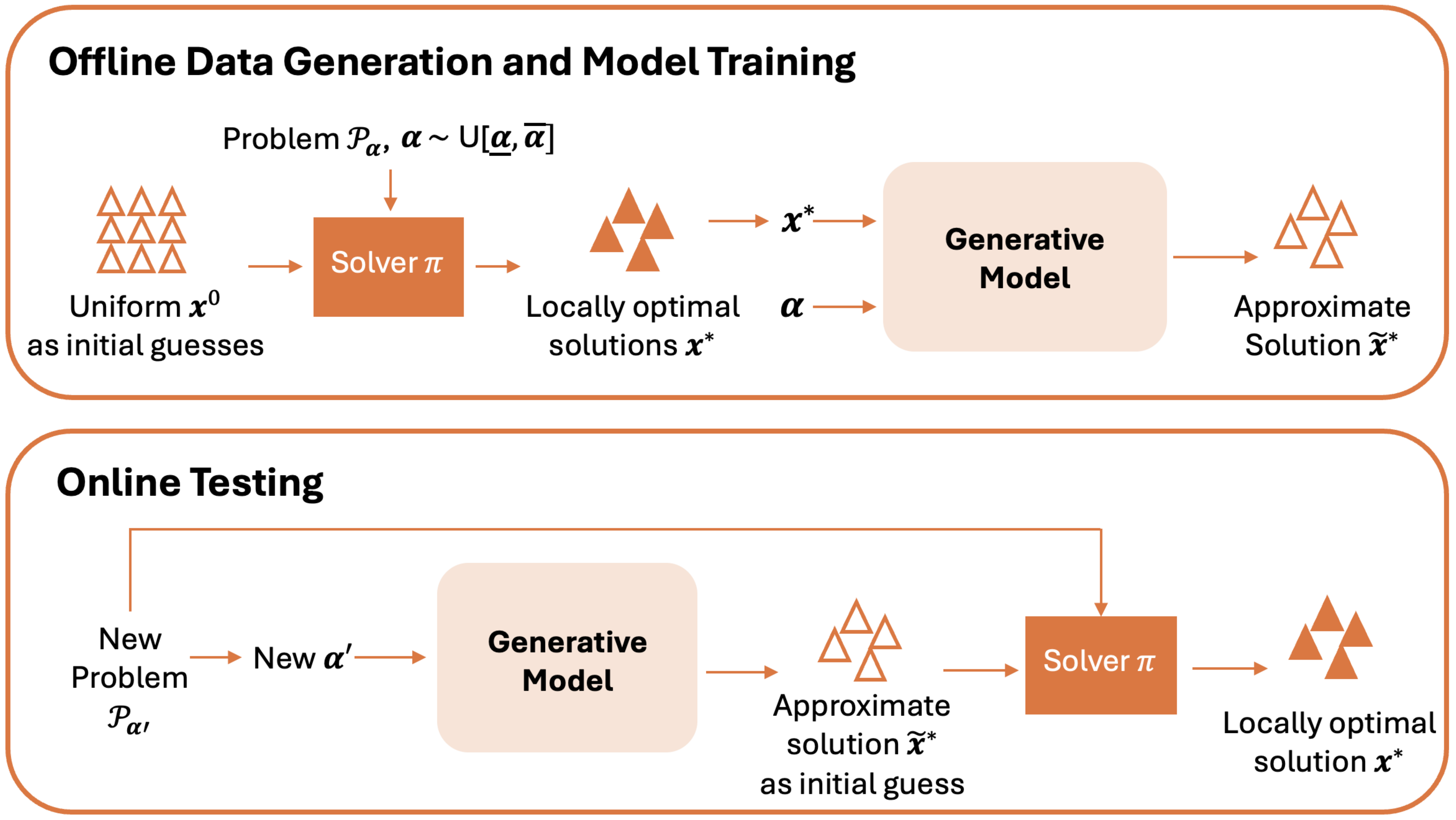}
    \caption{Workflow of the Amortized Global Search (AmorGS) Framework.}
    \label{fig:amorgs}
\end{figure}

The AmorGS framework aims to improve the computational efficiency of the global search for optimal control problems in a data-driven fashion, while maintaining the optimality and feasibility guarantee of the solutions.
The entire workflow of AmorGS is illustrated in Figure \ref{fig:amorgs}.
Offline, we adopt a numerical optimization solver $\pi$ to solve many optimal control problems with uniformly sampled parameter $\alpha$ and initial guesses $x^0$.
Then we collect a set of locally optimal solutions $x^*$ and filter out sub-optimal solutions to form a training dataset.
A conditional generative model is utilized to sample approximate solutions conditioned on the parameter $\alpha$.
In the online testing phase, when a new problem with unseen parameter value $\alpha'$ is presented, the generative model is able to predict the corresponding solution.
The solutions will be used as initial guesses for the solver $\pi$ to derive the final locally optimal and feasible solutions.

The key insight is that, the generative model can capture the solution structure in the training data and how the structure varies with parameter $\alpha$.
Thus the generative model can generalize to unseen problems and sample the structured solutions that are close to the groundtruth, from which the solver will quickly converge.

\subsection{Diffusion Probabilistic Models}

The diffusion probabilistic model is the state-of-the-art generative model that is capable of modeling and sampling variables with complex distribution, and it features a stable training process \cite{sohl2015deep, song2020score, ho2020denoising}.
This model comprises a forward diffusion process and a reverse process.
In the forward diffusion process, it incrementally adds small amounts of Gaussian noise to the training data $x_0 \sim p(x)$ for $T$ sampling steps, until the data approximates a Gaussian distribution.
The process is guided with a variance schedule $\{\beta_i\}_{i=1}^T$.
Consequently, the forward process generates a sequence of gradually corrupted data $x_0, x_1, x_2, ..., x_T$:
\begin{align} \label{eq: forward diffusion process}
    q(x_t | x_{t-1}) &= \mathcal{N}(x_t; \sqrt{1 - \beta_t} x_{t-1}, \beta_t \mathbf{I}) \nonumber \\
    x_t &= \sqrt{\bar \alpha_t} x_0 + \sqrt{1 - \bar \alpha_t} \epsilon, \quad t = 1, 2, ..., T
\end{align}
where $\alpha_t = 1 - \beta_t$ and $\bar \alpha_t = \prod_{i=1}^t \alpha_i$, and $\epsilon \sim \mathcal{N}(0, \mathbf{I})$.
We use noises with increasing variance in the forward process, thus $\beta_1 < \beta_2 < ... < \beta_T$.

In the reverse process, we learn how to denoise the data at each sampling step by learning the corresponding Gaussian mean $\mu_\theta$ and variance $\Sigma_\theta$ with a neural network $\theta$:
\begin{align}
    p_\theta(x_{t-1}|x_t) = \mathcal{N}(x_{t-1}; \mu_\theta(x_t, t), \Sigma_\theta(x_t, t)), \quad t = 1, 2, ..., T
\end{align}
As shown in Denoising Diffusion Probabilistic Model (DDPM) \cite{ho2020denoising}, more efficient training can be achieved by parameterizing $\epsilon_\theta(x_t, t)$ instead, which is to predict the noise $\epsilon$ added on $x_{t-1}$ to obtain $x_t$.
The loss function to train $\epsilon_\theta(x_t, t)$ is as follows:
\begin{align}
    \mathcal{L} = \mathbb{E}_{x_0, t, \epsilon} || \epsilon_\theta(x_t(x_0, \epsilon), t) - \epsilon ||_2^2
\end{align}
where $\epsilon \sim \mathcal{N}(0, \mathbf{I})$, and $t$ uniformly sampled from $\{1, 2, ..., T\}$.
Therefore, with $x_T$ sampled from $\mathcal{N}(0, \mathbf{I})$ and $\epsilon_\theta(x_t, t)$ for each sampling step $t$, we can sample the original data $x_0$ with distribution $p(x)$.

For the conditional sampling of $x \sim p(x|y)$ with $y$ being the conditional variable, we introduce $y$ in the reverse process $p_\theta(x_{t-1}|x_t, y)$.
We choose the classifier-free guidance method \cite{ho2022classifier} to further enhance the condition information, which jointly learns a conditional noise prediction $\epsilon_\theta(x_t, t, y)$ and an unconditional noise prediction $\epsilon_\theta(x_t, t, y=\emptyset)$, where $y$ is discarded by some probability $p_{uncond}$.
These two noise predictions can be learned using one neural network with the following loss function:
\begin{align}
    \mathcal{L}_{condition} = \mathbb{E}_{(x_0, y), t, \epsilon, b} || \epsilon_\theta(x_t(x_0, \epsilon), t, (1-b) \cdot y + b \cdot \emptyset) - \epsilon ||_2^2, \nonumber 
\end{align}
where $\epsilon \sim \mathcal{N}(0, \mathbf{I})$, $b \sim \text{Bernoulli}(p_{uncond})$, and $t$ uniformly sampled from $\{1, 2, ..., T\}$.

During the sampling, we use $\bar \epsilon_\theta(x_t, t, y)$ in the reverse process at each sampling step as the convex combination of the conditional noise $\epsilon_\theta(x_t, t, y)$ and unconditional noise $\epsilon_\theta(x_t, t, y = \emptyset)$, where $w$ is used to balance the sample quality and diversity.
\begin{align}
    \bar \epsilon_\theta(x_t, t, y) = \omega \cdot \epsilon_\theta(x_t, t, y) + (1 - \omega) \cdot \epsilon_\theta(x_t, t, y = \emptyset)
\end{align}

%% file: sections/results.tex
\section{Results}
\subsection{Example Problem I: Hybrid Cost Function}
\subsubsection{Data Generation}
In this problem, the conditional parameter $\alpha=\omega$ represents the weighting of the cost function in Eq. \eqref{equation: hybrid cost function}. 
We aim to observe how the decision vector changes with differently weighted objectives and train a diffusion model to predict this shift for unseen values of $\alpha$.
The data used to train the diffusion model comprises $165,000$ total feasible solutions for $11$ fixed values of $\alpha$, ranging from 0.0 to 1.0 in increments of 0.1.
We generate this data by using uniformly sampled decision vectors as initial guesses and running SNOPT in optimal mode with a maximum solving time of $240$ s.
The generated solutions are then filtered based on their objective values, retaining only the best $80\,\%$.
\subsubsection{Solving Time Study}
The quality of the samples produced by the diffusion model is tested by using them to warm-start SNOPT and comparing the runtimes and solution ratios to those from uniformly sampled initial guesses. 
We conduct this evaluation for $\alpha = 0.05$, $\alpha = 0.15$ and $\alpha = 0.95$, values not included in the dataset used for training. 
These test values are chosen because they are near the boundaries of the $\alpha\in[0,1]$ interval and within ranges where significant changes occur, as demonstrated in the following section.
Consequently, they present the greatest challenge for generating good initial guesses. 
The feasibility and optimality ratios of this comparison, based on 1,000 initializations per $\alpha$ value and method, are shown in Table \ref{tab:methods_comparison_hybrid} for a maximum solving time of $240\,\mathrm{s}$.

Both the feasibility ratio and the ratio of locally optimal solutions are significantly higher when using initial guesses sampled from the diffusion model.
No locally optimal solutions are generated within the maximum solving time when using uniform sampling, highlighting the problem’s complexity.
Notably, the diffusion model’s ability to generate locally optimal solutions within the maximum solving time is remarkable, considering that only feasible solutions were used for training. 
This suggests that a two-step process could be effective: initially generating a large number of locally optimal solutions with the diffusion model, followed by training a second, improved model exclusively on these locally optimal solutions.

The mean solving time for the diffusion model is only slightly lower than that for uniform sampling, as the values in Table \ref{tab:methods_comparison_hybrid} show.
This is because the majority of initial guesses that produce no solutions, or solutions that are feasible but not locally optimal, hit the maximum solving time of $240\,\mathrm{s}$ due to the fact that we run SNOPT in optimal mode. 
The solving times are based on runs conducted on 2.8 GHz Intel Cascade Lake processors and include, in the case of the diffusion model, a sampling time of $77\,\mathrm{s}$ for generating the $1,000$ initializations on an NVIDIA A100 GPU.
This corresponds to an additional sampling time of $0.077\,\mathrm{s}$ per solution, although this duration depends on the number of initializations sampled at once.

\begin{table}[!t]
    \centering
    \begin{tabular}{llcccc}
        \toprule
        \textbf{$\alpha$} & \textbf{Method} & \textbf{Feasibility Ratio} & \textbf{Optimality Ratio} & \textbf{Time (Mean±STD)}[s] \\
        \midrule
        \multirow{2}{*}{0.05} & Uniform Sampling & 30.6 \% & 0.0 \% & 239.8 ± 4.4 \\
                             & Diffusion Model & \textbf{47.4} \% & \textbf{10.6 \%} & \textbf{222.5 ± 38.8}\\
        \midrule
        \multirow{2}{*}{0.15} & Uniform Sampling & 27.8 \% & 0.0 \% & 239.7 ± 6.6 \\
                             & Diffusion Model & \textbf{53.5} \% & \textbf{15.4} \% & \textbf{217.0 ± 44.8}\\
        \midrule
        \multirow{2}{*}{0.95} & Uniform Sampling & 35.1 \% & 0.0 \% & 240.1 ± 2.8 \\
                             & Diffusion Model & \textbf{46.4} \% & \textbf{5.7} \% & \textbf{229.7 ± 33.4}\\
        \bottomrule
    \end{tabular}
    \caption{Comparison of feasibility and optimality ratio and (sampling + solving) time statistics for the feasible solutions from 1,000 samples with 3 different weights of the hybrid cost function, that were not included in the training data. The limit on the maximum solving time is $240\,\mathrm{s}$.}
    \label{tab:methods_comparison_hybrid}
\end{table}

\subsubsection{Solution Structure}
After demonstrating the AmorGS framework’s ability to accelerate the global search and produce feasible and locally optimal solutions at a high rate, we examine how the distribution of solutions changes with varying $\alpha$.
In particular, we show that the diffusion model can capture structure in the solution space and generalize to unseen weights of the hybrid cost function. 

\begin{figure}[!t]
    \centering
    \includegraphics[width=0.8\linewidth]{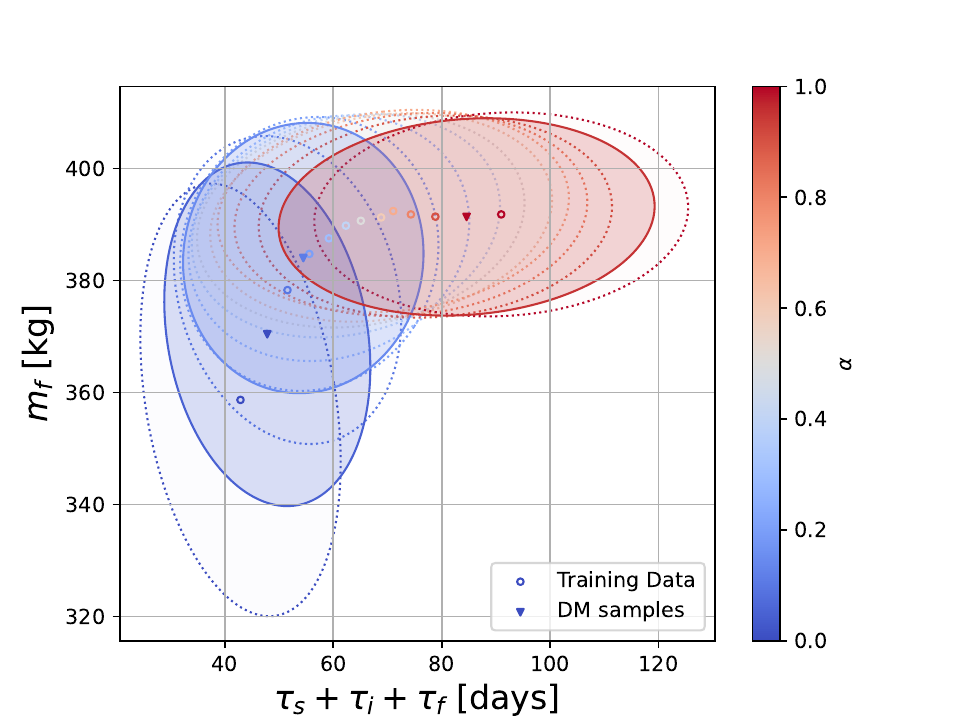}
    \caption{The average time of flight and final fuel mass are visualized for fixed objective function weights $\alpha$ from the training dataset. Each point is at the center of an ellipse, representing the corresponding covariance matrix based on one standard deviation. The samples from the diffusion model (DM) for three $\alpha$ values not included in the training data are shown as triangles, with the corresponding covariance ellipses highlighted.}
    \label{fig:TOF mf average}
\end{figure}

Fig. \ref{fig:TOF mf average} illustrates how the solution clusters shift in the $m_f$-time-of-flight plane when the objective function weight $\alpha$ changes.
The mean of all training solutions for a given $\alpha$ is represented as a circle, with the corresponding ellipse visualizing the covariance matrix based on one standard deviation to indicate the spread and correlation of the data around the mean. 
As expected, the solutions shift towards lower fuel consumption and longer time of flight with increasing $\alpha$.
Beyond $\alpha = 0.5$, there is no substantial improvement in average fuel consumption, while the time of flight continues to rise.

The diffusion model correctly predicts this shift in the distribution, as demonstrated by the location of the sample means between the neighboring points in the training data. 
The mean and covariance ellipses for the unseen values of $\alpha = 0.05$, $\alpha = 0.15$ and $\alpha = 0.95$ correspond to 10,000 solutions sampled from the diffusion model.
These $\alpha$ values were chosen for testing, because the most significant changes in the distributions occur within those ranges. 

\begin{figure}[!b]
    \centering
    \includegraphics[width=0.99\linewidth]{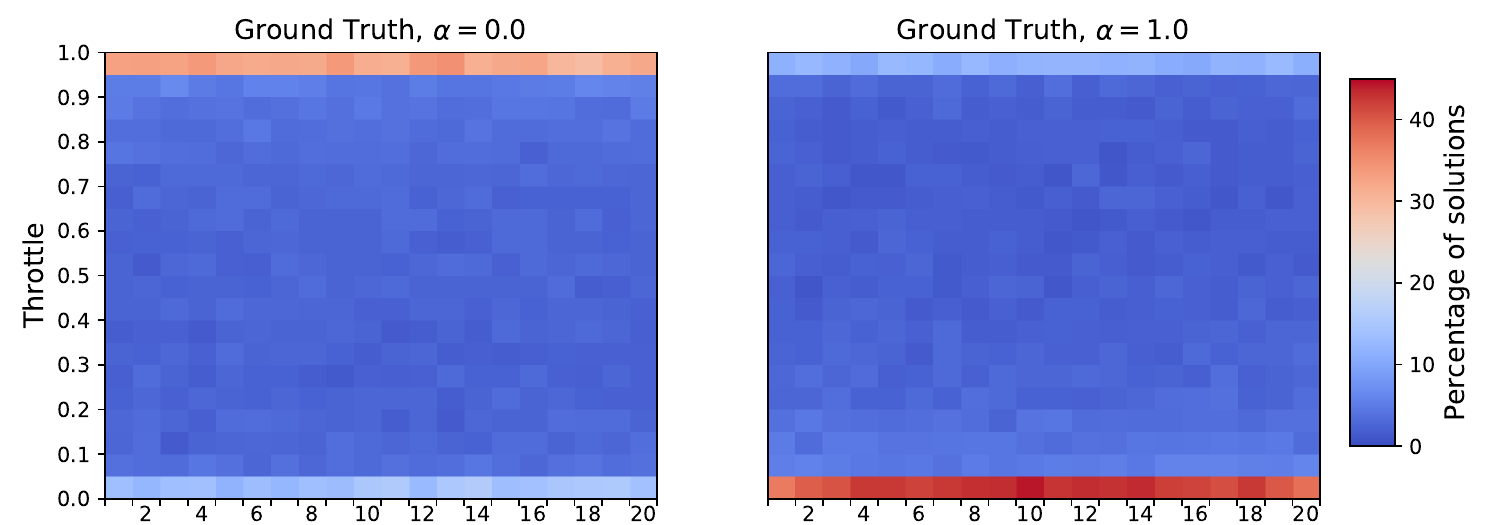}
    \caption{The throttle profiles for the minimum time ($\alpha=0$) and minimum fuel ($\alpha=1$) cost function are shown as a color map, visualizing the density of 1,500 solutions from the training data. The range of the throttle is split into fixed intervals, and the percentage of solutions within each interval is shown as a colored rectangle for each segment.}
    \label{fig: throttle plots min time and fuel}
\end{figure}
The diffusion model is able to generate new solutions for different objective functions by generalizing the learned distribution to unseen values of the conditional variable. 
The similar shape and size of the covariance ellipses in comparison to the neighbouring ellipses demonstrates that not only is the mean of the data reasonable, but the distribution and spread can also be predicted, proving the model’s ability to generate diverse samples.

\begin{figure}[!b]
    \centering
    \includegraphics[width=0.99\linewidth]{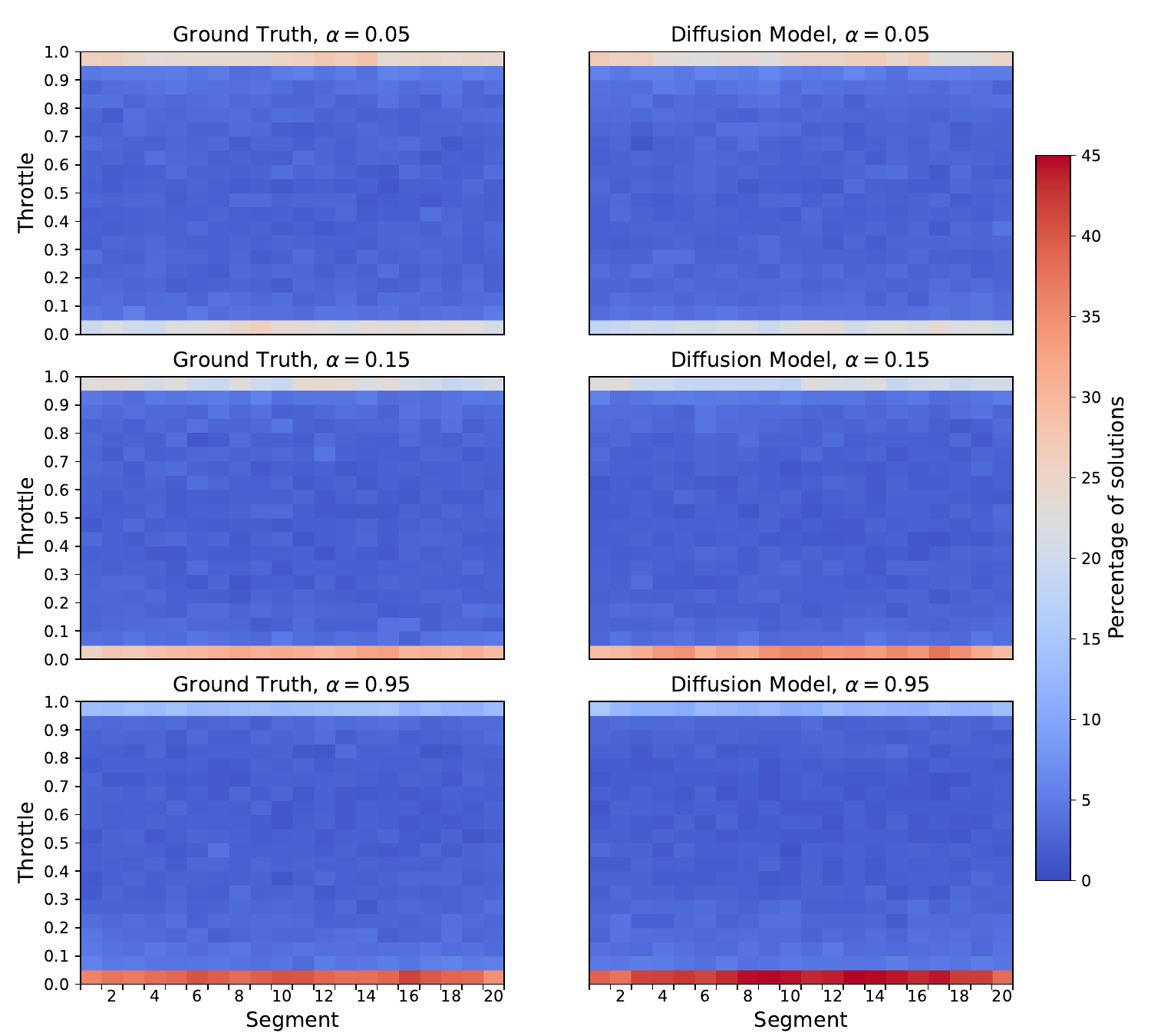}
    \caption{The throttle profiles for different levels of the cost function weight $\alpha$ are shown as a color map, visualizing the density of solutions. The range of the throttle is split into fixed intervals, and the percentage of solutions within each interval is shown as a colored rectangle for each segment. The ground truth data consists of 1,500 separately generated control solutions per $\alpha$ using uniform sampling, and the diffusion model data corresponds to 1,500 samples from the model.}
    \label{fig: throttle color plots}
\end{figure}
The results in Fig. \ref{fig:TOF mf average} demonstrate that the diffusion model generates accurate predictions for the decision variables directly related to the objective function. 
As the objective shifts from minimizing time to minimizing fuel, we also expect to see a corresponding shift in the throttle magnitude during each segment, transitioning from continuous thrusting solutions to typical ‘bang-bang’ behavior. 
Although we use a limited number of thrust segments ($N=20$), which cannot perfectly replicate this behavior, a clear trend is observed in Fig. \ref{fig: throttle plots min time and fuel}.
This figure shows the distribution of throttle values in each segment for $\alpha=0$ and $\alpha=1$.
The colormap indicates the percentage of solutions within a fixed throttle interval for each segment. 
In the minimum time case with $\alpha=0.0$, a high fraction of throttle values is close to $1.0$, indicating a trend towards full-thrust.
Conversely, in the minimum fuel case with $\alpha=1$, the majority of solutions have throttle values close to 0.0 in most segments. 
However, there is still an increased percentage of solutions with throttle values close to 1.0 compared to other throttle levels, suggesting a tendency towards ‘bang-bang’.
In both cases, the presence of solutions with smaller throttle values in some segments reveals that not all solutions in the training dataset exhibit perfect full-thrust  or 'bang-bang' behavior.

The diffusion model correctly predicts the shift from minimum time to minimum fuel as the comparison for the cost function weights of $\alpha=0.05$, $\alpha=0.15$ and $\alpha=0.95$ in Fig. \ref{fig: throttle color plots} demonstrates.
The distribution for samples from the model is displayed alongside separately generated ground truth solutions that were not included in the training set.
As $\alpha$ increases, the model’s samples show a decrease in the number of segments with full throttle and a corresponding increase in segments where the thrusters are turned off. The percentage of samples with maximum and minimum throttle closely matches the corresponding ground truth solutions.
\subsection{Example Problem II: Variable Terminal Boundary Condition}
\subsubsection{Data Generation}
\begin{figure}[!htbp]
    \centering
    \includegraphics[width=0.9\linewidth]{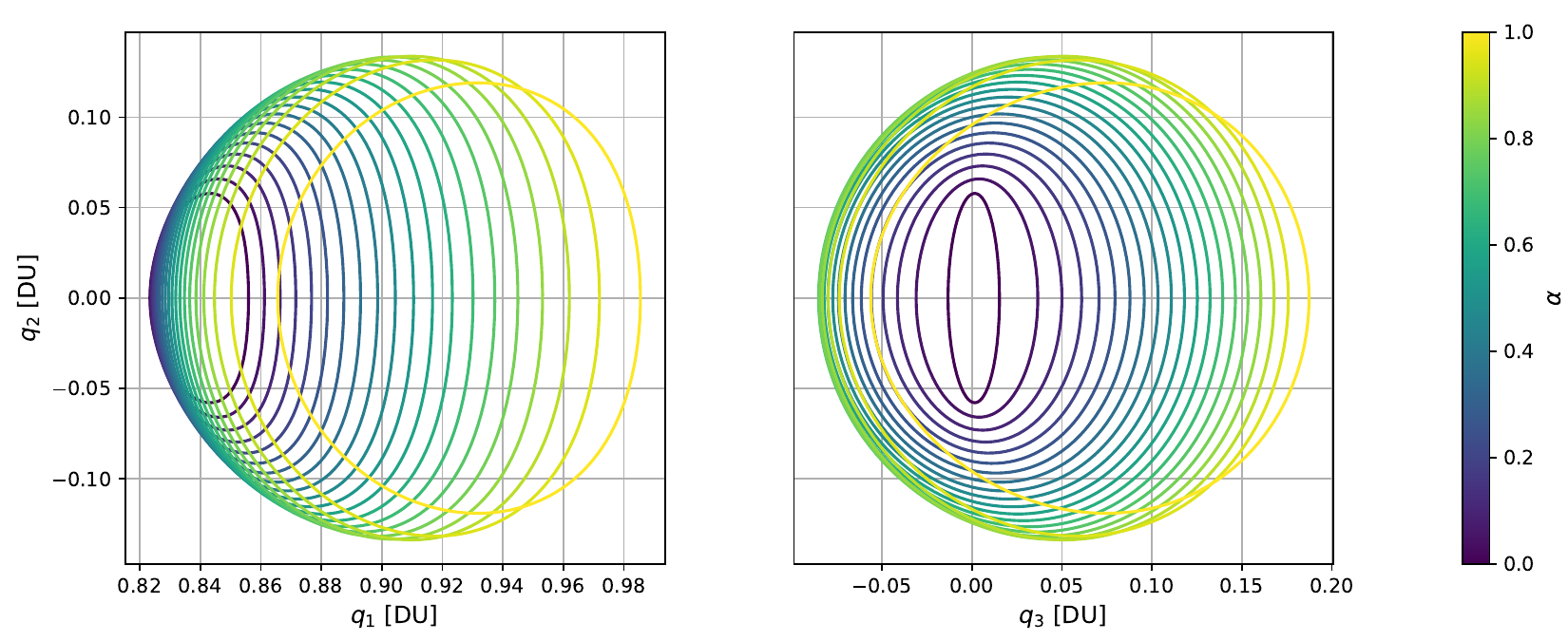}
    \caption{Halo orbits with varying energy levels in the range $\alpha \in [0,1]$, displayed in the $q_1,q_2$-plane and $q_3,q_2$-plane of the CR3BP coordinate system.}
    \label{fig:Halo orbits}
\end{figure}

For the second problem we aim to learn how the solution structure, particularly the decision variables $t_1$ and $t_2$, changes with variations in the orbital energy of the target halo orbit. 
The range of the perturbation energy in Eq. \eqref{equation: halo energy} is selected as $e_{pert,min} = 0.008$ and $e_{pert,max} = 0.095$ in natural units.
In this range, we can solve for an unambiguously defined set of halo orbits, visualized in Fig. \ref{fig:Halo orbits}.
To train the diffusion model, a large dataset of locally optimal control solutions across varying energy levels is necessary. 
We explore the entire solution space by uniformly sampling the decision vector from Eq. \eqref{equation: extended decision vector} and using these samples to warm-start the NLP solver SNOPT in optimal mode.
The parameter $\alpha$ is also sampled from a uniform distribution. 
We filter the generated locally optimal solutions based on the objective value, retaining only the $90\,\%$ of solutions with the highest final mass for training. 
The final training dataset comprises $100,000$ solutions, evenly distributed across all $\alpha$ values.
\subsubsection{Solving Time Study}
We evaluate the diffusion model’s ability to capture structure in the locally optimal control by generating initial guesses for a priori unseen energy levels $\alpha$ and warm-starting the solver.
Table \ref{tab:methods_comparison} compares different metrics for 1,000 runs, each warm-started from a uniformly sampled decision vector and a decision vector sampled from the diffusion model.
\begin{table}[!t]
    \centering
    \begin{tabular}{lcccc}
        \toprule
        \textbf{Method} & \textbf{Optimal Ratio} & \textbf{Mean(±STD)} [s] & \textbf{25\%-Quantile} [s] & \textbf{Median} [s] \\
        \midrule
        Uniform Sampling & 57.2 \% & 186.65 ± 120.58 & 88.42 & 124.26 \\
        Diffusion Model & \textbf{81.9} \% & \textbf{65.95 ± 86.02} & \textbf{12.14} & \textbf{28.49} \\
        \bottomrule
    \end{tabular}
    \caption{Comparison of optimality ratio and (sampling + solving) time statistics for the locally optimal solutions from 1,000 samples of the same set of random $\alpha\in[0,1]$.}
    \label{tab:methods_comparison}
\end{table}
Both methods use the same randomly sampled $\alpha$ values, which were not part of the model’s training data.
The diffusion model’s high-quality initial guesses are evident through a significantly higher ratio of locally optimal solutions within the maximum solving time of $500\,\mathrm{s}$.

\begin{figure}[!b]
    \centering
    \begin{minipage}[t]{0.48\textwidth}
        \centering
        \includegraphics[width=\textwidth]{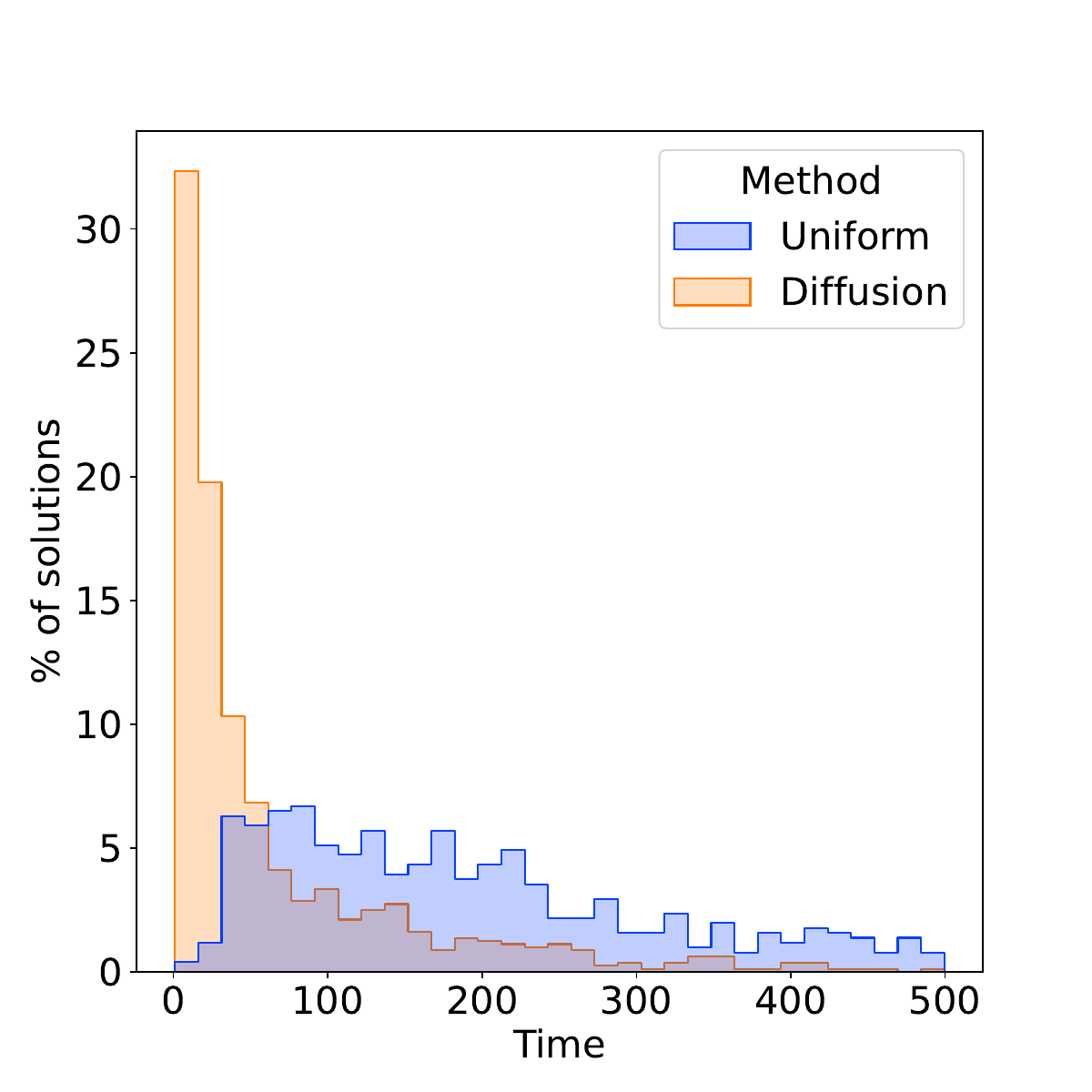}
        \caption{Histogram of solving times for locally optimal solutions with initializations sampled from a uniform distribution and the diffusion model. The solutions of 1,000 initializations for the same randomly chosen energy levels $\alpha\in[0,1]$ are shown for both methods.}
        \label{fig:Time Histogram}
    \end{minipage}
    \hfill
    \begin{minipage}[t]{0.48\textwidth}
        \centering
        \includegraphics[width=\textwidth]{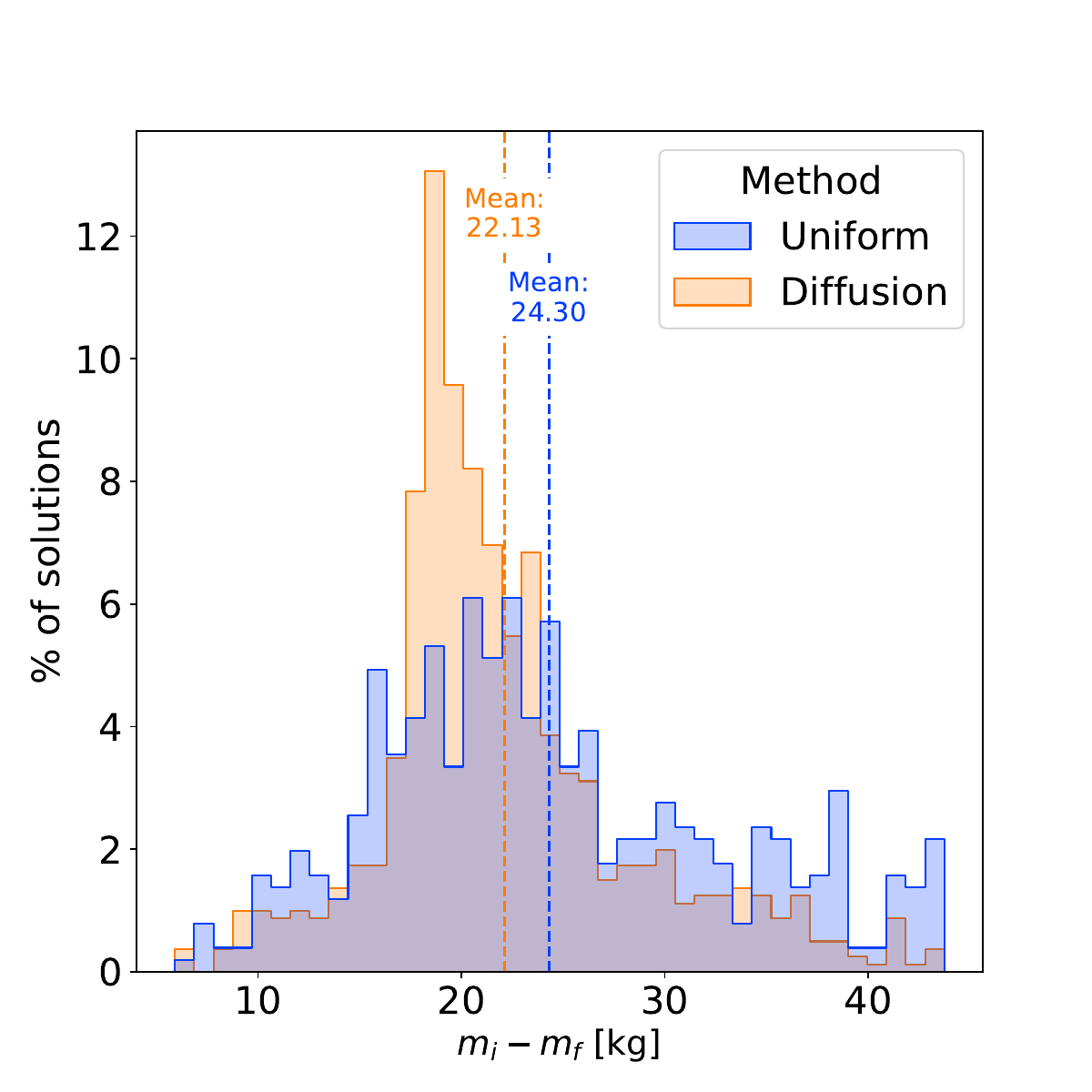}
        \caption{Histogram and mean of consumed fuel mass for locally optimal solutions with initializations sampled from a uniform distribution and the diffusion model. The solutions of 1,000 initializations for the same randomly chosen energy levels $\alpha\in[0,1]$ are shown for both methods.}
        \label{fig:fuel mass histogram}
    \end{minipage}
\end{figure}

The mean solving time of locally optimal solutions initialized with diffusion model samples is $2.83$ times lower than for uniformly sampled initial guesses.
The shorter average solving time is explained by the time histogram in Fig. \ref{fig:Time Histogram}, which shows the percentage of locally optimal solutions that were generated within certain ranges of solver runtimes.
The solving time statistics are based on runs conducted on 2.8 GHz Intel Cascade Lake processors and include, in the case of the diffusion model, a sampling time of $75\,\mathrm{s}$ for generating the $1,000$ initializations on an NVIDIA A100 GPU.
This results in an additional sampling time of $0.075\,\mathrm{s}$ per solution, although this duration depends on the number of initializations sampled at once.     

Fig. \ref{fig:Time Histogram} highlights that the advantage of the diffusion model in terms of optimality ration is even greater when using a shorter solving time.
The diffusion model’s strength lies in its ability to provide initial guesses that quickly converge to local optima, with $25\,\%$ of solutions being found within $12.08\,\mathrm{s}$, a performance 7.28 times better than that of uniformly sampled initial guesses. 
This demonstrates that combining the diffusion model with a numerical solver allows generating solutions at a very high rate by using a short maximum solving time.

The histogram of the consumed fuel mass for the same set of solutions in Fig. \ref{fig:fuel mass histogram} demonstrates that the initializations from the diffusion model generate solutions that span the entire range of objective values in the dataset, with a slightly lower mean compared to solutions obtained from uniformly sampled initial guesses.
The region around $m_i-m_f = 20\,\mathrm{kg}$, where many solutions are generated through uniform sampling is amplified to a strong peak in the solutions from the model.
This occurs because the diffusion model samples regions with high data density more frequently. Consequently, there are slightly fewer solutions generated at the edges of the interval.

\begin{figure}[!b]
    \centering
    \begin{subfigure}[b]{0.5\textwidth}
        \centering
        \includegraphics[width=\linewidth]{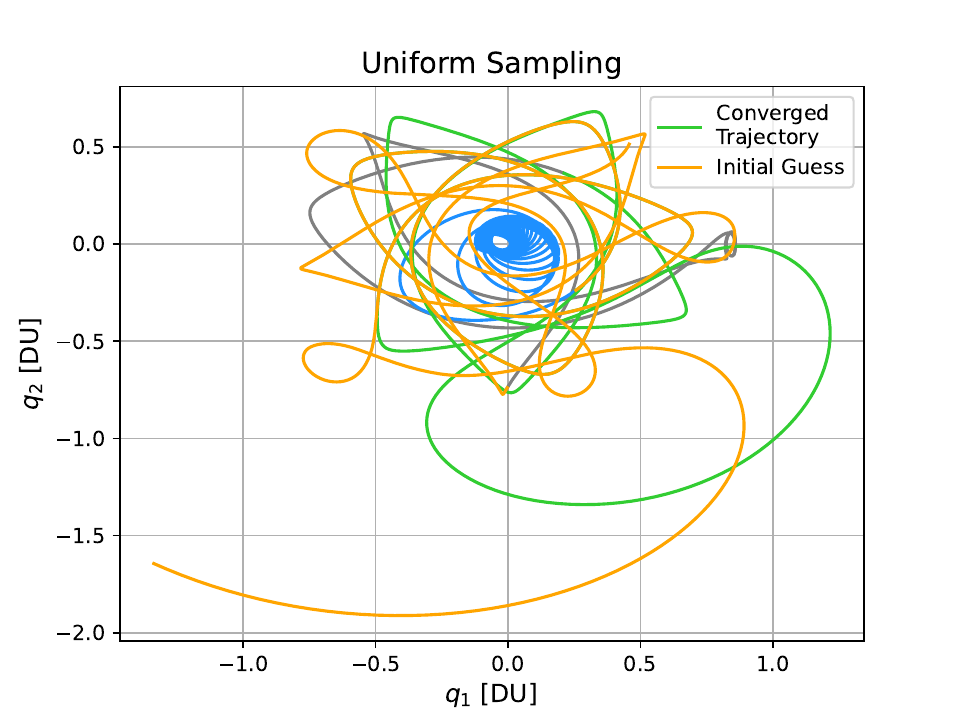}
    \end{subfigure}\hfill
    \begin{subfigure}[b]{0.5\textwidth}
        \centering
        \includegraphics[width=\linewidth]{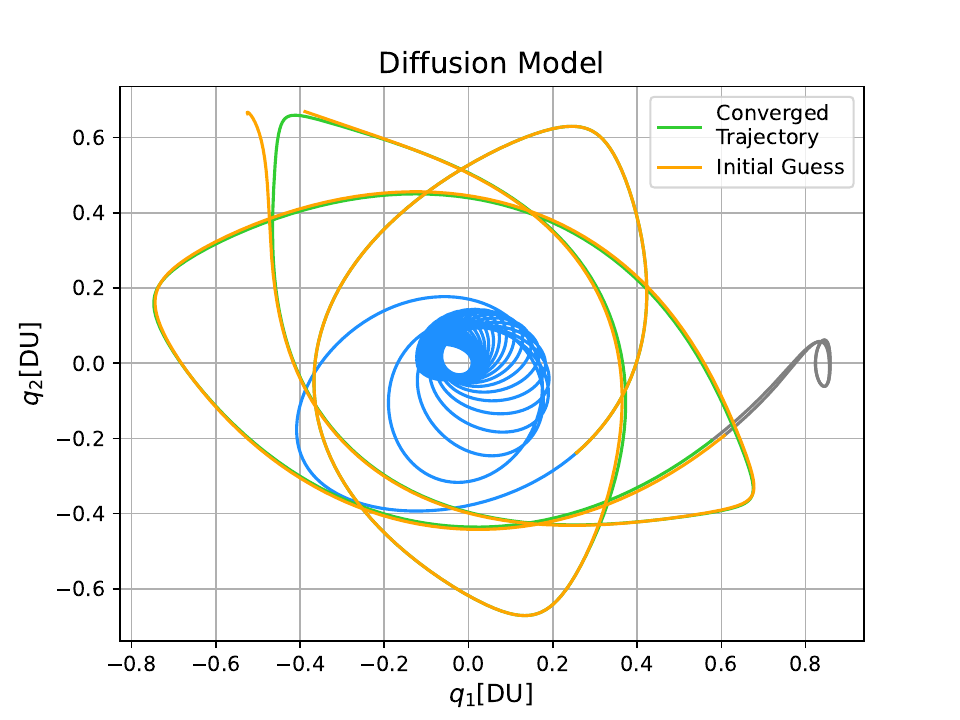}
    \end{subfigure}
    \caption{Visualization of the trajectories corresponding to an initial control guess (orange) and the final locally optimal trajectories (green) with a uniformly sampled control vector on the left and a sample from the diffusion model on the right. Both solutions correspond to an energy level of $\alpha = 0.024$. The trajectories start from the endpoint of a GTO spiral (blue) and end on a halo manifold arc (grey).}
    \label{fig:trajectory comparison}
\end{figure}

The shorter average solving times for the diffusion model can be explained by visualizing the trajectories corresponding to initial guesses for the two methods and comparing them to the final locally optimal solutions from the solver. 
Fig. \ref{fig:trajectory comparison} illustrates this for a randomly chosen energy level of $\alpha = 0.024$, with the uniformly sampled initial guess on the left and the sample from the diffusion model on the right. 
Using a forward-backward shooting approach, constraint violations for the initial guesses can be observed halfway along the trajectories.
For the uniformly sampled initial guess, this gap in the middle of the trajectory is large, and the initial trajectory appears fundamentally different from the locally optimal solution to which the solver converged. 
Both the time of flight and the controls $t_1$ and $t_2$, which define the manifold arc, change significantly during optimization, resulting in a solving time of $412.5\,\mathrm{s}$.
In contrast, the control vector sampled from the diffusion model results in a trajectory with a smaller constraint violation, visible as a gap in the top left of the image. 
The solver closes this gap without major alterations to most of the variables in the decision vector, leading to a shorter solving time of $17.9\,\mathrm{s}$.

\subsubsection{Solution Structure}
\begin{figure}[!b]
    \centering
    {
        \includegraphics[width=0.45\textwidth]{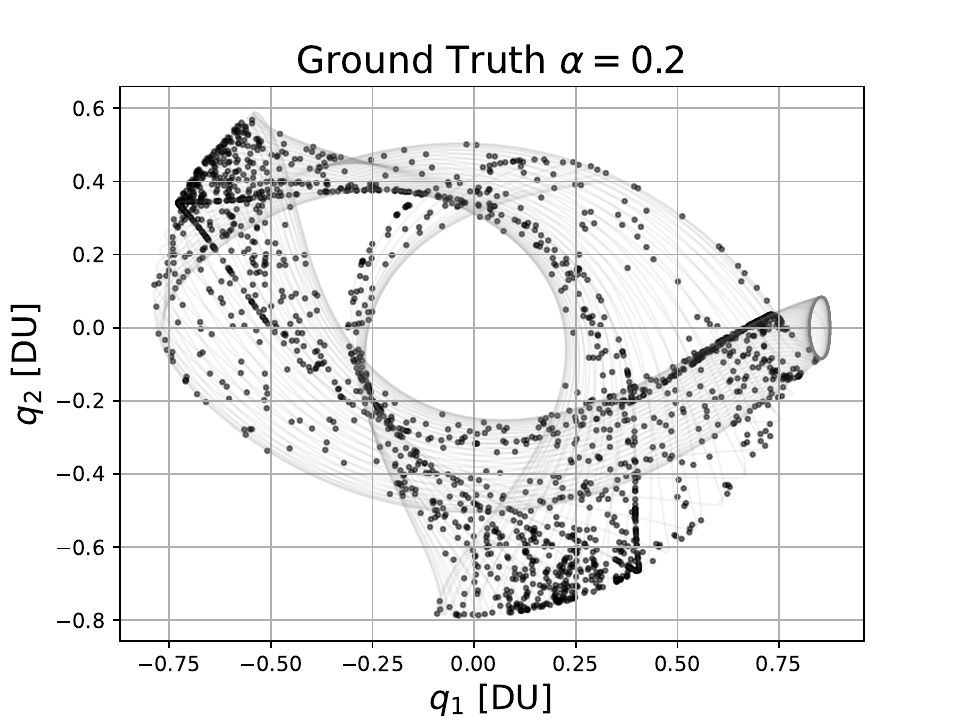}
    }
    {
        \includegraphics[width=0.45\textwidth]{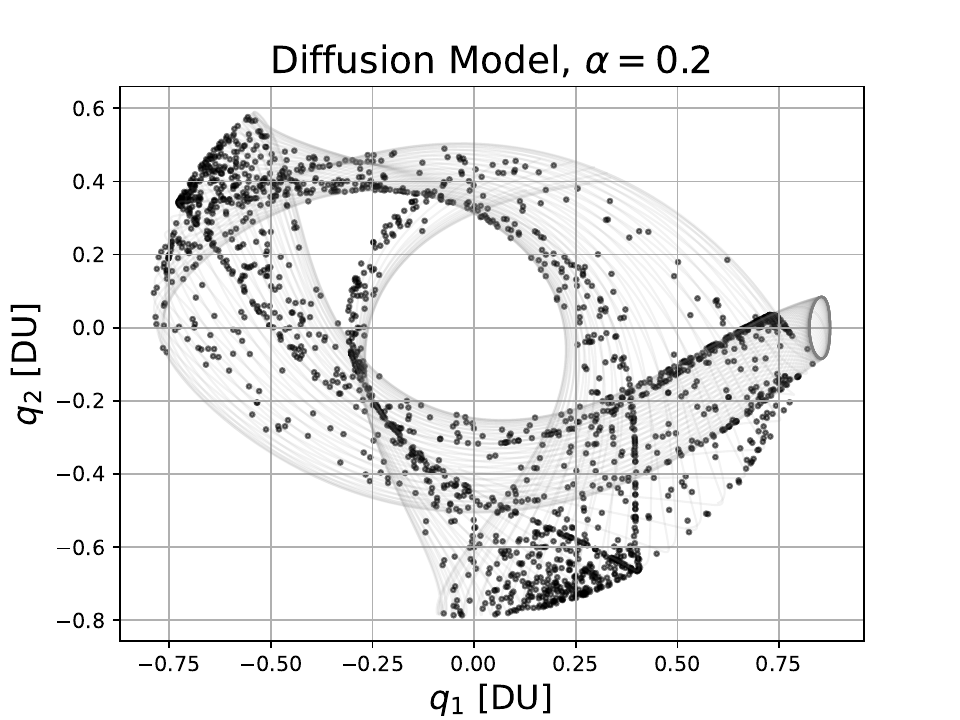}
    }
    \\
    {
        \includegraphics[width=0.45\textwidth]{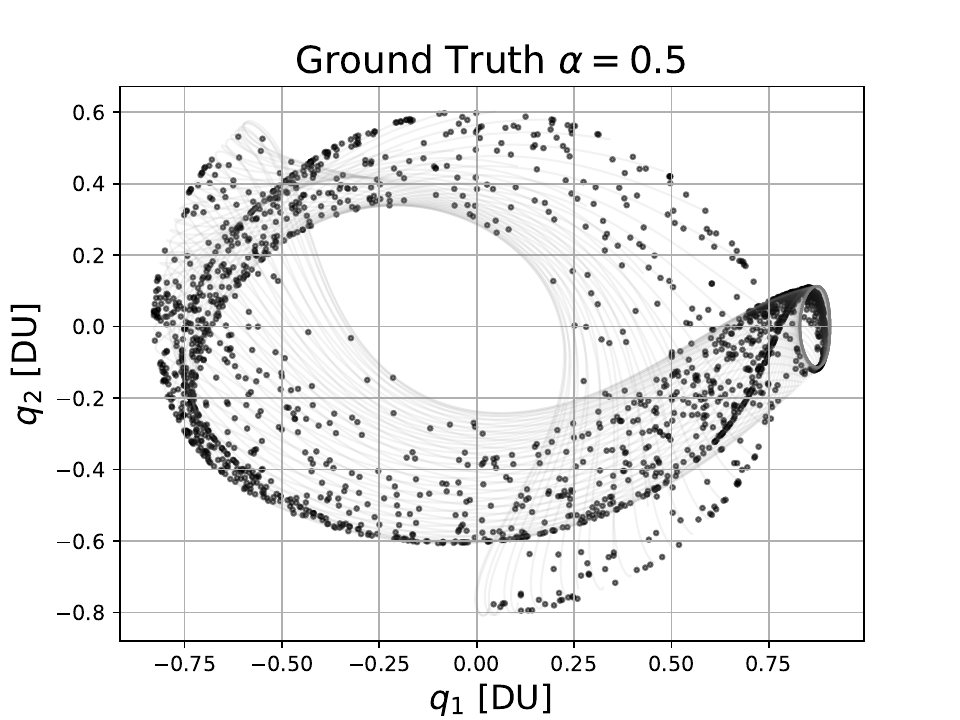}
    }
    {
        \includegraphics[width=0.45\textwidth]{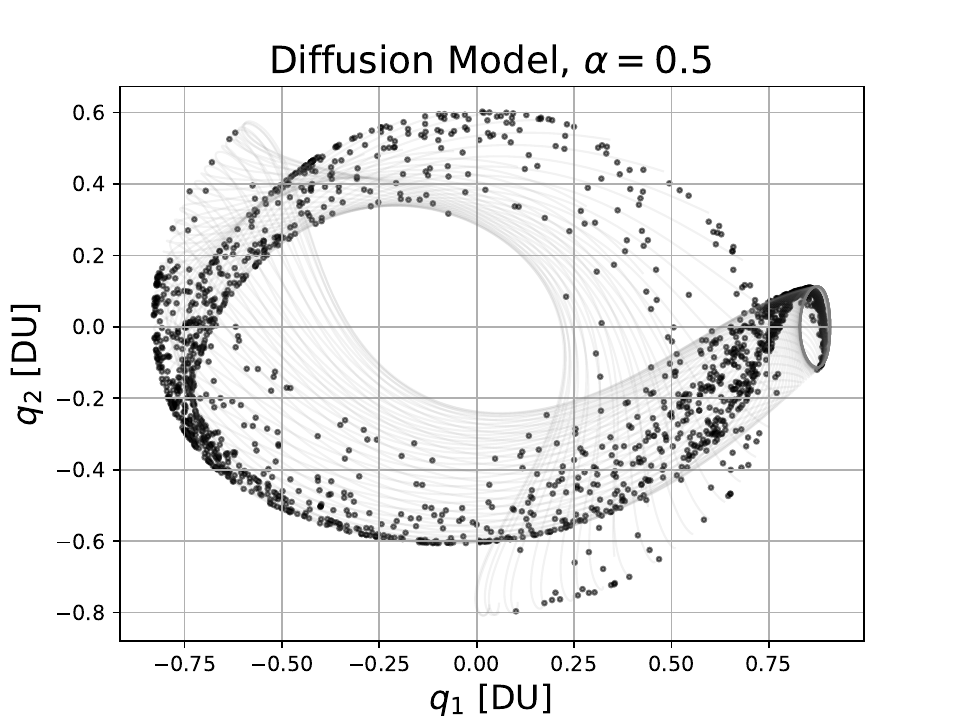}
    }
    \\
    {
        \includegraphics[width=0.45\textwidth]{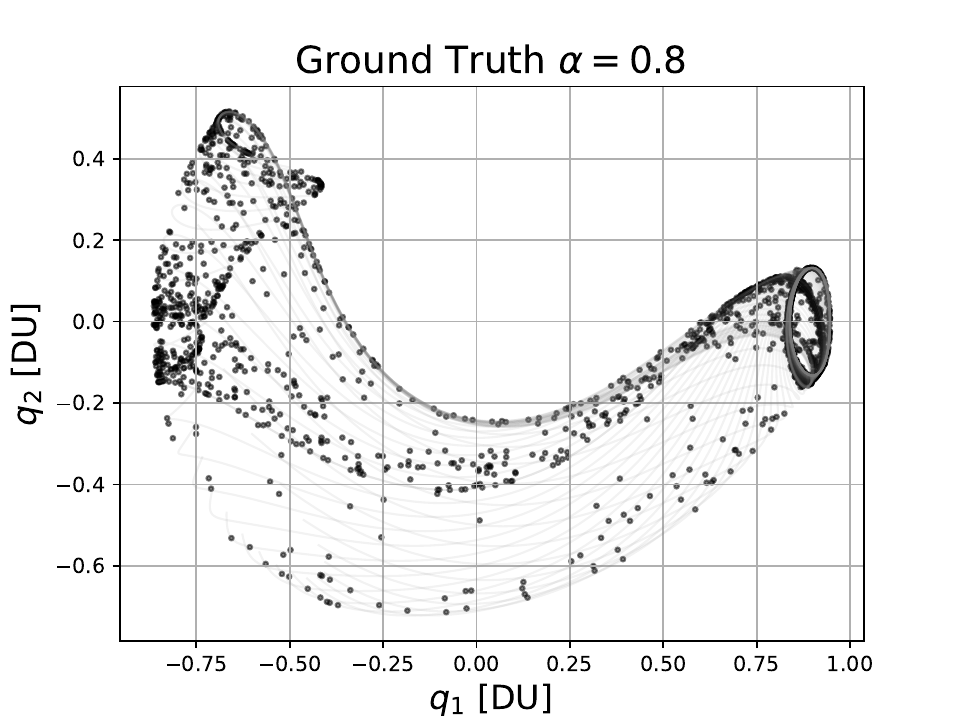}
    }
    {
        \includegraphics[width=0.45\textwidth]{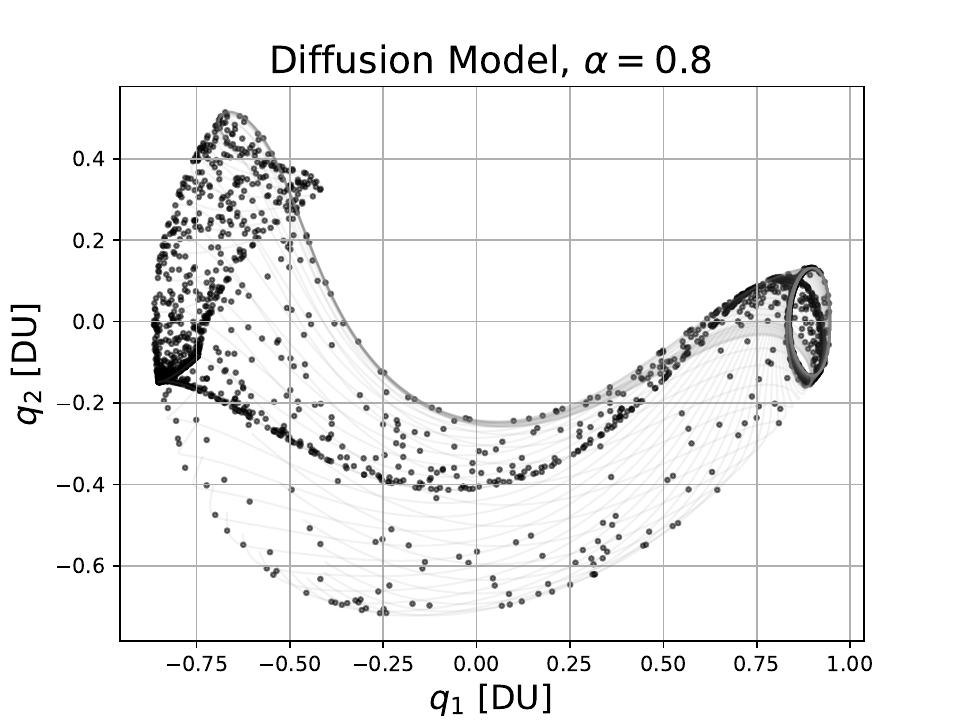}
    }
    \caption{Visualizations of the trajectory endpoints on manifold arcs for different solutions as projections onto the $q_1,q_2$-plane of the CR3BP frame, with the invariant manifold in the background. On the left, the ground truth data is shown for 2,000 solutions for three different energy levels. On the right, 2,000 samples produced by the Diffusion Model are shown for the same energy levels.}
    \label{fig:background manifold plots}
\end{figure}

The main focus of this problem is to see how the distribution of the temporal parameters $t_1$ and $t_2$, which characterize the invariant manifold used to coast onto the halo orbit, changes with varying energy levels $\alpha$.
In the left column of Fig. \ref{fig:background manifold plots}, the shift in the structure of these two variables is shown for three fixed energy levels as a projection onto the $q_1,q_2$-plane.
The background visualizes the stable, left invariant manifold of the corresponding halo orbit for the fixed bounds on $t_1$ and $t_2$. 
As the energy level increases, the length of the manifold decreases for a fixed $t_{2,max}$.
Each plot displays 2,000 solutions not used for training, generated using uniform sampling and SNOPT. 
The solutions are represented by the endpoint of their trajectory on the manifold.
The decision variable $t_2$ determines which specific manifold arc the endpoint is on, while $t_1$ defines the exact location on that arc where the trajectory ends. 
There are clusters on the manifold that represent regions on the $(t_1 \times t_2)$ grid that minimize fuel usage.  
These regions likely correspond to basins of attraction for local minima in the solution space.
The shape and size of the manifold, as well as the location of these clusters, shifts as the energy level increases.

The diffusion model is able to predict how the location of these clusters changes for different $\alpha$, without prior knowledge of the specific energy level’s structure, as the right column of Fig. \ref{fig:background manifold plots} demonstrates.
The comparison shows the distribution captured by the model for the same energy levels as the ground truth data on the left. 
Each point represents a $t_1,t_2$ pair corresponding to a decision vector sampled from the model for the chosen conditional parameter $\alpha$.

The model successfully captures the structure in the decision variables representing the terminal boundary condition and generalizes across changing halo energies. 
Therefore, a sample from the diffusion model has a high likelihood of being located in one of the basins of the solution space and converge quickly to the corresponding local minimum when used as an initial guess for the solver.
The spread of sampled initial guesses across large areas of the $(t_1 \times t_2)$ domain demonstrates the model’s ability to generate diverse initial guesses.

%% file: sections/conclusion.tex
\section{Conclusion}
In this paper, we demonstrate that a diffusion model can effectively capture structure in control solutions for trajectory optimization problems based on changing problem parameters. 
Li et al. \cite{li2023amortized, li2024efficient} previously showed that generative models can accelerate the global search for low-thrust transfers in the CR3BP space and predict hyperplane structures in time variables for different maximum thrust levels. 
Building on this foundation, we showcase that the ability to capture structure in the solutions space extends to additional entries of the decision vector and different conditional variables. 

We show that the model can be conditioned on the objective function itself, enabling rapid adaption of the weighting of mission goals such as fuel consumption and minimum flight time during the mission design phase. 
By utilizing a hybrid cost function and training the diffusion model on fixed weights, we can quickly generate new solutions for objective functions with different weightings.
In addition to shooting and coasting times and the final mass, the diffusion model accurately predicts the structure in the throttle profile and how this structure changes for different objectives.
Leveraging this capability, the model generates samples that converge to feasibility and local optimality significantly faster than uniformly sampled initial guesses. 

With the second example problem, we demonstrate that the diffusion model can learn to utilize dynamical structures for mission design in the CR3BP space. By employing a variable terminal boundary condition, we reveal clusters in the solution space that represent optimal locations to coast onto the stable invariant manifold of a halo orbit. The positions of these clusters depend on the orbital energy of the halo orbit, and our model accurately predicts these distributions for unseen energy levels. This capability enables us to generate accurate approximations of control solutions for trajectories targeting a wide range of halo orbits with different energies.
By actively incorporating invariant manifolds into the mission design, our global search framework predicts optimal manifold arcs and the precise points to coast onto them.

Both examples further showcase the strength of generative models in the early stages of space mission design.
Sampling from a diffusion model enables the efficient generation of new solutions and offers high flexibility regarding changing mission goals and parameters.
One limitation of the diffusion model is the upfront cost associated with generating a large amount of training data and training the model.
In future work, we will explore methods to reduce these costs while also enhancing the model’s performance.
In addition, it will be interesting to study how the diffusion model generalizes to problems that slightly shift from the training data distribution, without introducing much computational overhead.